# MSLE: An ontology for Materials Science Laboratory Equipment – Large-Scale Devices for Materials Characterization

Mehrdad Jalali[1, 2], Matthias Mail[2], Rossella Aversa[3], and Christian Kübel[2, 4, 5]

[1]Institute of Functional Interfaces (IFG), Karlsruhe Institute of Technology (KIT), Hermann-von Helmholtz-Platz 1, 76344 Eggenstein-Leopoldshafen, Germany

[2]Institute of Nanotechnology (INT), Karlsruhe Institute of Technology (KIT), Hermann-von Helmholtz-Platz 1, 76344 Eggenstein-Leopoldshafen, Germany

[3]Steinbuch Centre for Computing (SCC), Karlsruhe Institute of Technology, Hermann-von-Helmholtz-Platz 1, 76344 Eggenstein-Leopoldshafen, Germany

[4]Helmholtz Institute Ulm for Electrochemical Energy Storage (HIU), Karlsruhe Institute of Technology, Hermann-von-Helmholtz-Platz 1, 76344 Eggenstein-Leopoldshafen, Germany

[5]Department of Materials and Earth Sciences, Technical University Darmstadt, Alarich-Weiss-Str. 2, 64287 Darmstadt, Germany

## Abstract

This paper introduces a new ontology for Materials Science Laboratory Equipment, termed MSLE. A fundamental issue with materials science laboratory (hereafter lab) equipment in the real world is that scientists work with various types of equipment with multiple specifications. For example, there are many electron microscopes with different parameters in chemical and physical labs. A critical development to unify the description is to build an equipment domain ontology as basic semantic knowledge and to guide the user to work with the equipment appropriately. Here, we propose to develop a consistent ontology for equipment, the MSLE ontology. In the MSLE, two main existing ontologies, the Semantic Sensor Network (SSN) and the Material Vocabulary (MatVoc), have been integrated into the MSLE core to build a coherent ontology. Since various acronyms and terms have been used for equipment, this paper proposes an approach to use a Simple Knowledge Organization System (SKOS) to represent the hierarchical structure of equipment terms. Equipment terms were collected in various languages and abbreviations and coded into the MSLE using the SKOS model. The ontology development was conducted in close collaboration with domain experts and focused on the large-scale devices for materials characterization available in our research group. Competency questions are expected to be addressed through the MSLE ontology. Constraints are modeled in the Shapes Query Language (SHACL); a prototype is shown and validated to show the value of the modeling constraints.

## 1. Introduction

An ontology can provide a formal description for a knowledge graph where the graph consists of a set of concepts within a domain and their relationships. For a more detailed explanation, it is necessary to formally specify components such as individuals (object instances), classes, attributes and relationships, restrictions, rules, and axioms.

Recent efforts have been made to develop ontologies and metadata in materials science. The European Materials Modeling Ontology (EMMO), the common standard for materials modeling, is developed by the European Materials Modelling Council (EMMC) [1]. Another recent approach has been adopted in Novel Materials Discovery (NOMAD)[2], a user-driven platform for sharing and exploiting computational materials science data. The Materials Design Ontology (MDO) [3] defines concepts and relationships to cover knowledge in the field of materials design (especially in solid-state physics). In [4], the authors proposed a research data infrastructure for materials science, Kadi4Mat, which extends and combines the features of an electronic lab notebook and a repository. All of this is aimed at maximizing the exchange of materials data generated by various researchers and organizations; for this reason, it is necessary to store the data in a way that is findable, accessible, interoperable, and reusable (FAIR).

Figure 1 is constructed using the VOSviewer software tool, showing a network visualization for the supplied keywords presented in the Scopus database articles by 2022. More specifically, the figures show the co-occurrence network and the topic clusters for the ontology and materials science keywords and their connections.

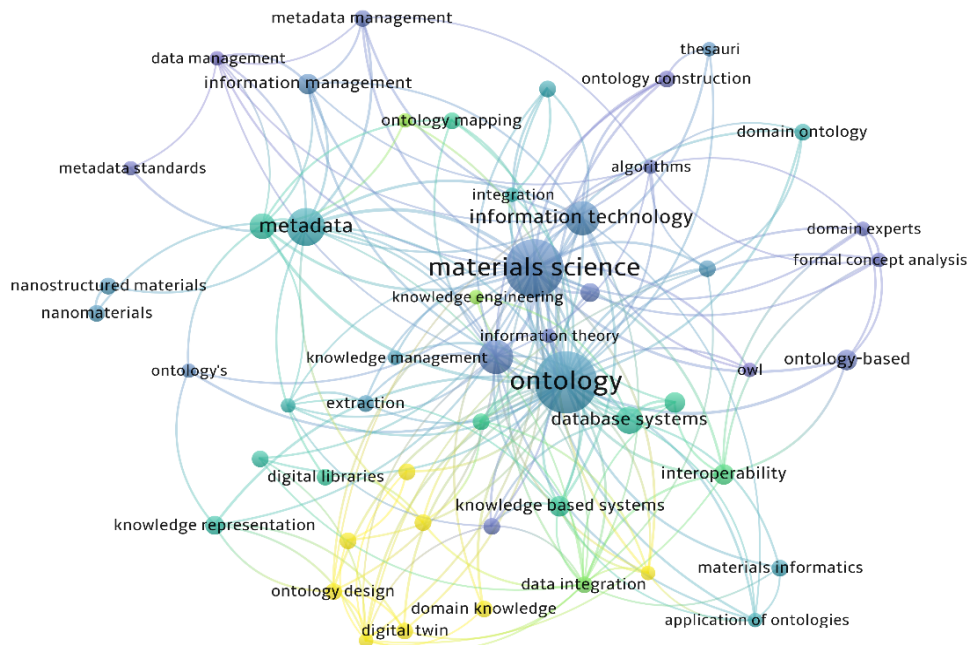

*Figure 1: Network connections for the ontology and materials science keywords with VOSviewer*

Many research institutes in materials science tend to have a large volume of research data and a large quantity of large-scale characterization lab equipment with different specifications. Consequently, it is challenging to plan experiments and analyses and to document and interpret results systematically. Thus, scientists can exploit an ontology focused on lab equipment, collecting the knowledge of device specifications and operating conditions to select devices and document parameter settings best suited for a given measurement (Figure 2). The MSLE presented in this paper addresses these challenges.

For example, a researcher tends to start with an experiment to analyze a sample with Zeiss Auriga 60. Through the ontology, the researcher realized that the Zeiss Auriga 60 is a dual-beam electron microscope with some specific parameters that need to be understood before starting. Operation voltage in terms of high tension SEM and high tension FIB with a specific voltage range, imaging detectors such as STEM detector, In-lens, SE, and 4WBSD, Gas Injection System (GIS), imaging, analyzing, and sample preparation techniques such as STEM imaging, EBSD analysis, 3D Slice, and TEM sample preparation are such kind of parameters.

A detailed description of the ontology is provided in Section 2, and the implementation in Protégé is shown in Section 3.

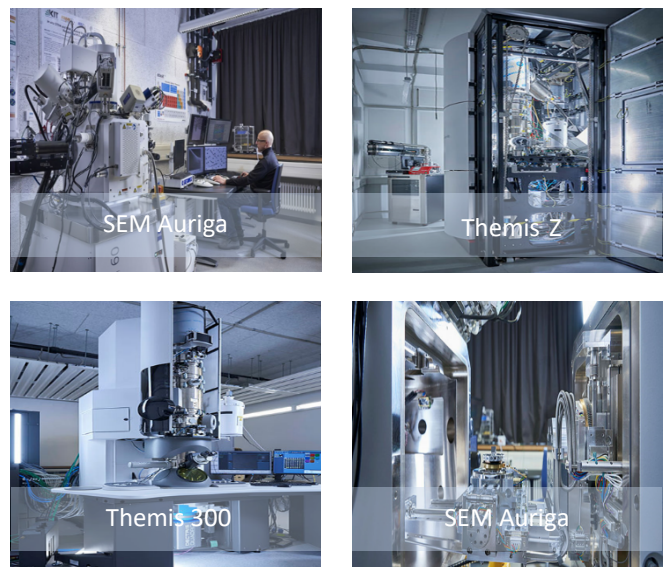

*Figure 2. It is challenging to plan experiments and analyses and consistently document and interpret the results of a large amount of large-scale laboratory characterization equipment with different specifications.*

## 2. Development of the MSLE ontology

The development of an ontology implies a sequence of steps to be performed: term definitions, community involvement, development of a common vocabulary, formalization of a common taxonomy, lightweight ontology, ontology alignment and validation, and heavyweight ontology. In this

Section, we describe each of them in detail, showing how we applied it to the specific case of the MSLE ontology. A workflow of model architecture used for ontology generation is shown in Figure 3.

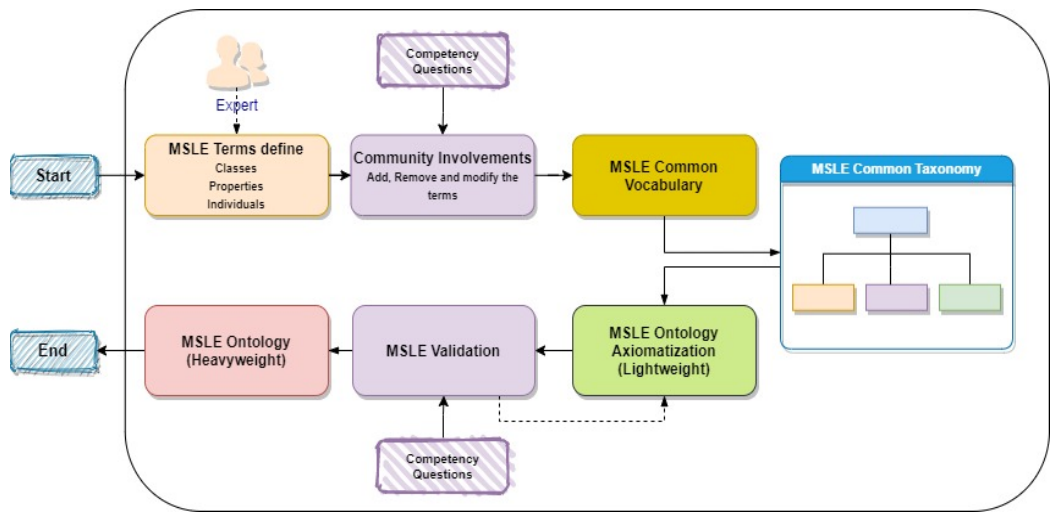

*Figure 3. Workflow of the proposed MSLE ontology generation.*

## 2.1 Term definitions in the MSLE

In this step, an unstructured list of all applicable terms expected to appear in the ontology has been collected with the support of domain experts or semi-automatic Natural language processing (NLP) through materials science lab equipment documents. An example of defined lab equipment terms can be found in Table 1. Therefore, ontology classes, properties, and instances can be defined by looking at all associated resources. In addition, other languages, such as German for experts familiar with only German names of equipment, may be attributed to the significant concepts of MSLE ontology.

*Table 1: Example of definition of some terms in materials lab equipment.*

| Terms | Term type in ontology | Description |
|---|---|---|
| Electron Microscope | Class | An electron microscope is a microscope that uses a beam of accelerated electrons as a source of illumination for analyzing materials. |
| Focused Ion Beam Microscope (FIB) | Class | A focused ion beam microscope (FIB) is an instrument using a focused ion beam for site-specific analysis, deposition, and ablation of materials. |
| Scanning Electron Microscope (SEM) | Class | A scanning electron microscope (SEM) is a microscope that uses a focused accelerated electron beam to analyze the surface of a material |
| Transmission Electron Microscope (TEM) | Class | A transmission electron microscope (TEM) is a microscope that uses an accelerated electron beam transmitted through a sample to analyze material. |
| Optional Equipment | Class | Additional equipment or devices could be installed on the primary equipment. |
| SEM Stage | Class | The stage is part of the microscope. It supports the sample and can be used to move it around. |
| High Tension | Property | The high tension refers to the potential difference used to accelerate the electron or ion beam in an electron or ion microscope after emission from the electron/ion gun. . |
| Operate with | Property | A property that mentions, for instance, that equipment works with some specific software. |
| Magnification | Property | The magnification of a microscope refers to the enlargement of an observed object in imaging. |
| Dimension??? | Property | This may be used to demonstrate sample dimensions. |
| Value | Property | A property to give a value for a device setting. |

## 2.2 Community Involvement in the MSLE

At this stage, to improve the semantics of the ontology, all terms extracted from the previous step were considered by a community of domain experts to add, delete or modify terms. This step is necessary to define and agree within the community on the terms' nomenclature and definition. Community ontologies are generally designed for broader community use, and it would be beneficial to have broad community involvement throughout the whole process of developing an ontology. Furthermore, at this stage, the scope of ontology can be clarified by reviewing the list of jurisdictional issues provided by the community of experts. Competency questions (CQs) are user-oriented interrogatives that allow us to scope our ontology [4]. In other words, they are some questions to which the users would like answers, exploring and questioning the ontology and its associated knowledge base. Competency questions specify what knowledge has to be entailed in the ontology and thus can be seen as a set of requirements on the content and a way of scoping and delimiting the subject domain that has to be represented in the ontology. In practice, the competency questions used to determine the ontology limits include: Does the ontology contain enough information to answer these types of (competency) questions? Do the answers require a particular level of detail or representation of a specific area? Table 2 provides some examples of competency questions.

*Table 2. An example of competency questions (all CQs based on the equipment available in certain institutions)*

| Competency Questions |
|---|
| What types of electron microscopes are available? |
| For which type of instrument is a gas injection system optional equipment? |
| Which kinds of spectrometers are available? |
| Can an EDX system be installed on an SEM? |
| Can an EBSD system be installed on an SEM? |
| Which detectors in SEM are available? |
| What is the range of high tensions for TEM? |
| What is the optional equipment for FIB/SEM? |
| What kind of emission sources are available for ion beam instruments? |

## 2.3 Development of common vocabulary in the MSLE

Among the advantages, ontology defines a common vocabulary for researchers who need to share information in a domain. At this stage, the MSLE vocabulary is defined as a set of concepts in a machine-readable format. The Web Ontology Language (OWL) [5], the Resource Description Framework (RDF)[6], and the Protégé software [7] have been used to develop the MSLE ontology. SKOS [8] provides a data model and vocabulary for expressing Knowledge Organization Systems (KOSs) such as thesauri and classification schemes in the Semantic Web. SKOS can annotate a concept for display or search purposes with specific properties for definition, preferred alternative, and hidden labels (skos:definition, skos:prefLabel, skos:altLabel, and skos:hiddenLabel). These properties are generally used for better and clearer expression of a concept in ontology. As an example, in Figure 4 skos: definition is used to define the scanning electron microscope concept, and skos:altlabel is used to define the alternative term for this concept using the well-known term "SEM" as well as the term in the other languages like the German "Rasterelektronenmikroskop". Furthermore, schema:image was used to link this concept to an image of this equipment, which may be helpful for beginners.

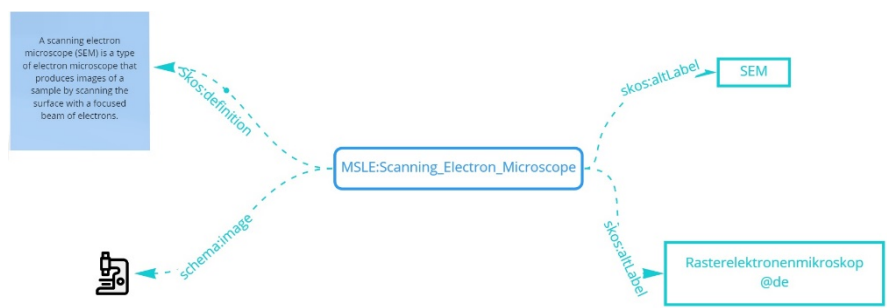

*Figure 4 Concept scheme for Scanning Electron Microscope equipment by using SKOS*

## 2.4 Formalization of common taxonomy in the MSLE

A taxonomy formalizes the hierarchical relationships among concepts and specifies the terms to refer to them [9]. In the MSLE, once the related concepts have been defined as a common vocabulary, the taxonomic relationships between them should be modeled hierarchically. Since the taxonomy is a form of the classification scheme, it can be derived by analyzing related research and documents in the domain and consulting experts using material science equipment [10, 11]. We can acquire more semantic knowledge in this field by defining a taxonomy. Figure 5 illustrates a portion of MSLE taxonomy.

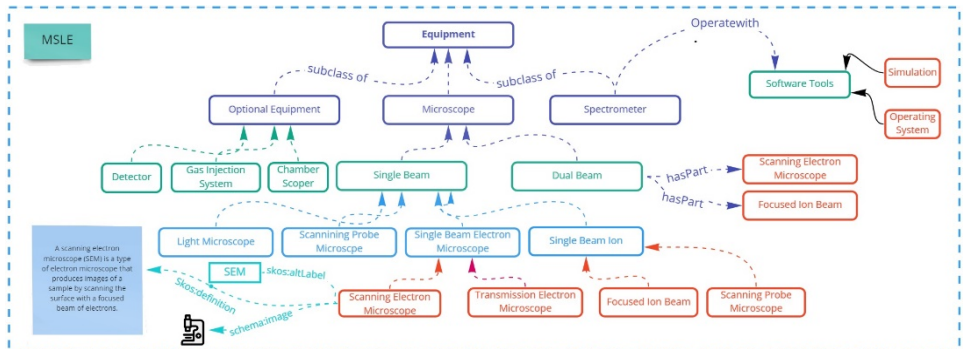

*Figure 5 a simplified conceptual model of the relations between equipment*

## 2.5 Lightweight ontology of MSLE

In the lightweight ontology step, the axioms, properties, and indeed general rules are defined in a formal language. For instance, "Has some Scanning_Coils SubClassOf Scanning_Electron_Microscope", "Dual beam equivalent to hasPart some ( Scanning_Electron_Microscopy and Focused_Ion_Beam) ".

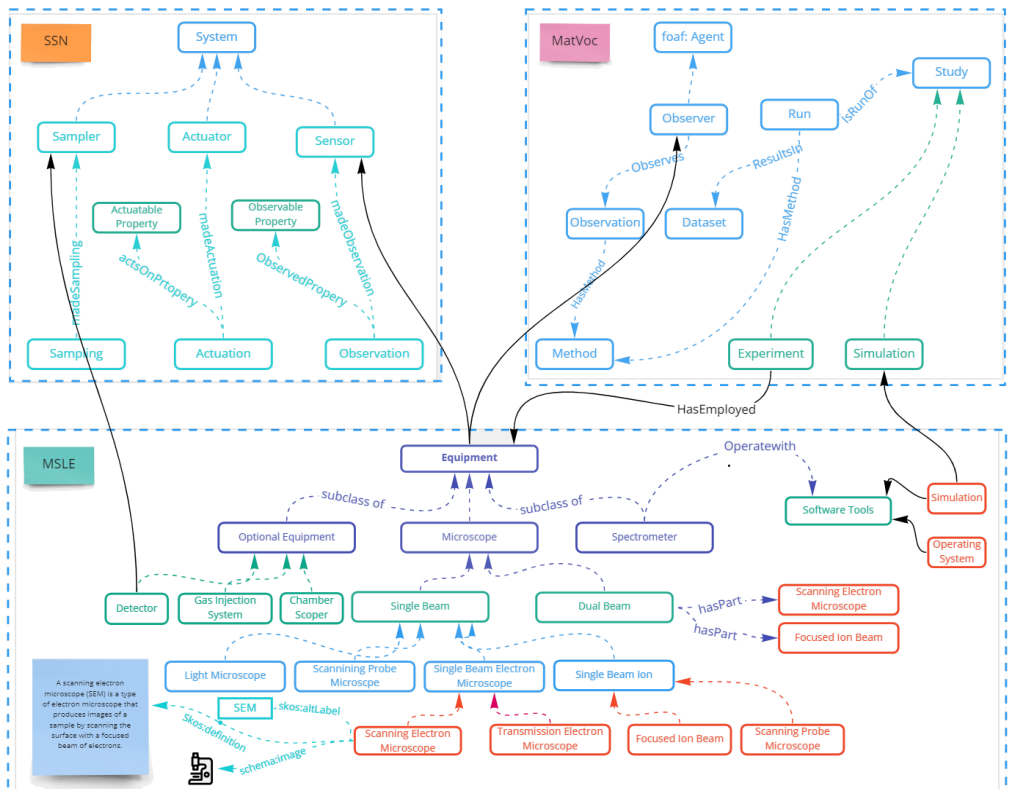

*Figure 6 MSLE ontology with integration with SSN and MatVoc ontologies*

Another critical task is ontology alignment, the process of finding corresponding entities with the same intended meaning in different ontologies [12]. For a better standardization of the MSLE, the two well-known ontologies, SSN [13] and MatVoc [14] have been integrated into the MSLE core. The SSN ontology describes sensors and their observations, the involved procedures, the studied features of interest, the samples used to do so, the observed properties, and the actuators. Since lab equipment can be considered sensors that observe physical and chemical phenomena, aligning the MSLE to the SSN ontology can improve its semantics. For instance, the *equipment* class in the MSLE is a *sensor* class in the SSN, or the *detector* class in the MSLE is a *sampler* in the SSN.

On the other hand, MatVoc provides resources to define processes (simulation, experiment, etc.), observations (measurement, method, etc.), and metadata (attributes, classification, etc.). Thus, it can be used to align experiment, simulation, observer, and metadata related to equipment used in the materials science domain to appropriate concepts in the MSLE. For instance, the *equipment* class in the MSLE is an *observer* in MatVoc or each *experiment* in MatVoc *HasEmployed equipment* in MSLE. Figure 6 illustrates an ontology integration based on MSLE, MatVoc, and SSN.

## 2.6 Maturity Model for ontology validation in the MSLE

After the ontology has been developed, it needs to be evaluated and verified by comparing it to requirements. This requires defining a maturity model for our ontology. Maturity models are widely used in process improvement, even in ontology-generating processes. Users of an ontological maturity model should be confident that the evaluated processes' weaknesses can be found and that the most valuable changes are made. Therefore, the evaluation of maturity models is an important activity that highlights the strengths and weaknesses of a system [15]. The maturity model for ontology-generating is a measure that can indicate how reliable the ontology is and how effective it is at self-improvement. Furthermore, with the maturity model, an assessment of the progress against the objectives can be performed. In this work, we defined four solutions for the MSLE maturity model, as shown in Figure 7.

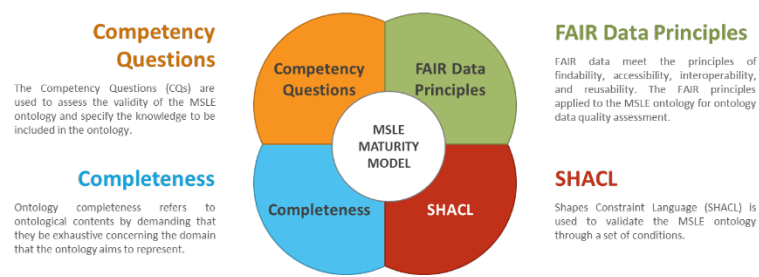

*Figure 7. Four solutions for the MSLE Maturity Model*

Both domain and ontology experts are actively involved in assessing the quality of an ontology using these solutions. These solutions include FAIR data principles, competency questions, shapes constraint language (SHACL), and ontology completeness.

*FAIR Data Principles*

To our knowledge, FAIR data meet the principles of findability, accessibility, interoperability, and reusability (FAIR)[16]. The FAIR data infrastructure is essential for making data findable and AI-ready in materials science research data [17]. Since the ontologies can be considered research data or metadata, publishing FAIR ontologies on the web is critical [18]. In addition to creating a federated semantic space to improve the global FAIRness of semantic artifacts, the created MSLE ontology should assess in FAIR aspect. Some standard protocols and tools should use to achieve this aim. The FAIRsFAIR project proposed preliminary recommendations related to the FAIR principles and best practice recommendations to improve the global FAIRness of semantic artifacts [19]. In this study, some indicators based on the FAIR principles are introduced to evaluate ontologies manually in terms of FAIRness. Moreover, software tools such as FAIRsharing [20] and F-UJI[21] are available to automatically assess standard FAIR indicators on the ontologies as metadata.

*Competency Questions*

Competency Questions (CQs) play an essential role in ontology development and evaluation. In the evaluation concept, this capability assists users in checking if CQs are being fulfilled by the ontology being defined. In the MSLE ontology, we used SPARQL [22], a semantic query language capable of retrieving and manipulating data stored in the Resource Description Framework (RDF) format for formulating the competency questions. Moreover, since the MSLE is a domain-specific ontology and ontology instances are inserted according to the laboratory equipment of each institution, the evaluation can be performed independently.

*Table 3* presents some competency questions and the related results on the MSLE ontology. The MSLE ontology can be evaluated using the percentage of accurate results. Moreover, since the MSLE is a domain-specific ontology and ontology instances are inserted according to the laboratory equipment of each institution, the evaluation can be performed independently.

*Table 3 Example of SPARQL for some Competency Questions on MSLE ontology (Examples mentioned in the third column are related to our Lab)*

| Competency Questions | SPARQL | Results |
|---|---|---|
| List of Single Beam Electron Microscopes | *SELECT ?SingleBeamEM*<br>*WHERE { ?SingleBeamEM rdfs:subClassOf MSLE:Single_Beam}* | Scanning_Electron_Microscope (SEM)<br>Single_Beam_Electron_Microscope<br>Transmission Electron Microscope (TEM)<br>… |
| What is the maximum high tension of the electron beam for Zeiss Auriga 60 | *SELECT ?High_Tension*<br>*WHERE { MSLE:Zeiss_Auriga_60 MSLE:hasHighTension ?High_Tension}* | 30 |
| What is a Dual Beam Microscope? | *SELECT ?X*<br>*WHERE { MSLE:Dual_Beam owl:equivalentClass ?X}* | (Explanation of the results: Any kind of microscope combining two different illumination sources.) |

| | | | Scanning_Electron_Microscope (SEM) and_ Focused_ion_beam (FIB) … |
|---|---|---|---|
| What types of detectors are available? | | SELECT ?Detectors<br>         WHERE { ?Detectors rdfs:subClassOf MSLE:Detectors} | 4QBSD_Detector<br>STEM_Detector<br>EsB<br>In_Lens_Detector<br>… |
| What is the range of SEM and FIB magnification for Zeiss Auriga 60? | | SELECT ?SEM_Magnification  ? FIB_Magnification<br>         WHERE { MSLE2:Zeiss_Auriga_60 MSLE:hasSEMmagnification  ?SEM_Magnification ;<br><br>MSLE:hasFIBmagnification  ?FIB_Magnification } | SEM_Magnification: 12 X – 1000 Kx<br>FIB_Magnification:300 x – 500 Kx |
| What types of the dual-beam microscope are available? | | SELECT ?DualBeam<br>         WHERE { ?DualBeam rdf:type MSLE:Dual_Beam} | Zeiss_Auriga_60<br>FEI_Strata_400s<br>… |
| In which dual beam system is the maximum high tension of the ion beam 30 kV? | | SELECT ?x<br>         WHERE { ?x rdf:type MSLE:Dual_Beam  ;<br>                      MSLE:hasHighTension "35" ^^xsd:integer<br>         } | Zeiss_Auriga_60<br>FEI_Strata_400s<br>… |
| Which instrument is equipped with a STEM detector? | | Select ?device<br>WHERE { ?Device  rdf:type ?x .<br>             ?x rdf:type  owl:Restriction .<br>             ?x  owl:onProperty MSLE:hasDetector .<br>             ?x  owl:someValuesFrom MSLE:STEM_Detector  } | Zeiss_Auriga_60<br>FEI_Strata_400s<br>… |
| List of all samplers in SSN ontology | | SELECT ?Detectors<br>WHERE {?Detectors rdfs:subClassOf SSN:Sampler } | STEM_Detector<br>SESI<br>4QBSD_Detector<br>Camera<br>Nothing<br>In-Lens_Detector<br>EsB |
| Which equipment has a gas injection system? | | SELECT ?Equipment<br>WHERE {?Equipment MSLEE:hasInjection ?X.<br>?X rdf:type MSLEE:Gas_Injection_System .<br>   } | FEI_Strata_400s |
| What are the types of FEI Strata 400S gas injection system (GIS)? | | SELECT ?GIS<br>WHERE {MSLEE:FEI_Strata_400s MSLEE:hasInjection ?GIS.<br>  } | Tungsten_W<br>Carbon_C<br>Platinum_Pt<br>XeF2 |

*Shapes Constraint Language (SHACL)*

In addition to the SPARQL approach, the quality assessment of the MSLE ontology can be performed using the Shapes Constraint Language (SHACL), a semantic language for validating RDF graphs against a set of conditions provided as shapes. RDF graphs used in this manner are called shapes graphs, and the RDF graphs validated against a shapes graph are called data graphs. OWL ontologies are good at describing the meaning of terms, but in contrast to RDF shapes, they have no mechanism to describe the type of things that those terms can express. SHACL can be used for validation, user interface building, code generation, and data integration [23]. Figure 8 shows an excerpt of a SHACL shape describing a dual-beam electron microscope. This example defines a shape :*MSLEShape* of type

*sh:NodeShape*. It has a target class declaration pointing to *:Dual_Beam,* which applies to all nodes that are instances of *:Dual_Beam.* The following lines declare that nodes confirming to *:MSLEShape* must satisfy the following constrains.

- They must have exactly one property *MSLE:hasHighTension* with values of type *xsd:integer*.
- They must have exactly one property *MSLE:hasHighTension* whose value must be in the range 0.1 and 30 kV.
- They must have at least one location as the value of property *MSLE:hasLocation* for dual beam.
- They must have at least one detector as the value of property *MSLE:hasDetector* for dual beam.

Figure 9 defines an RDF data graph corresponding to the shape graph. In this example, the ion beam of a Zeiss Auriga 60 is set to a high-tension value of 35. Since in shape, the high-tension value is set between 0.1 and 30 kV, the following error message will appear as below once the shape is executed on the data:

*sh:ResultMessage " The high tension for the dual beam needs to be in the proper range."*

```
@prefix sh:    <http://www.w3.org/ns/shacl#> .
@prefix xsd:   <http://www.w3.org/2001/XMLSchema#> .
@prefix MSLE:  <http://www.semanticweb.org/hr7456/ontologies/2021/8/MSLE#> .

MSLE:MSLEShape  a sh:NodeShape ;
   sh:targetClass MSLE:Dual_Beam ; # Applies to all SEM

  sh:property [
     sh:path MSLE:hasHighTension ;
     sh:minInclusive 0.1  ;
     sh:maxInclusive 30  ;
     sh:maxCount 1 ;
     sh:message " The high tension for the dual beam needs to be in the proper range. "@en ;
   ] ;

 sh:property [
     sh:path MSLE:hasHighTension ;
     sh:datatype xsd:integer ;
     sh:message " The data type of high tension needs to be Integer. "@en ;
   ];

sh:property [
     sh:path MSLE:hasDetector ;
     sh:minCount 1 ;
     sh:message "Needs to define a detector"@en ;
 ] ;

sh:property [
     sh:path MSLE:hasLocation ;
     sh:minCount 1 ;
     sh:message " The location for the equipment needs to be defined. "@en ;
 ] ;
```

*Figure 8 SHACL Constraint to ensure Dual Beam has the specific range and data type of High Tension.*

```
:Zeiss_Auriga_60 rdf:type :Dual_Beam ,
         owl:NamedIndividual ,
         [ rdf:type owl:Restriction ;
           owl:onProperty :hasParameter ;
           owl:someValuesFrom :HighTension
         ] ;

    :hasHighTension 35 ;
```

*Figure 9 RDF Data Graph for the high tension value of the Zeiss Auriga 60*

*Ontology Completeness*

Completeness measures the number of unique observations an ontology can make. It is subjective and mostly related to the scope of the ontology that defines based on the expert preferences in the domain. Completeness measures by assessing the percentage of accurate answers based on the competency questions for the related scope. It is evident an ontology might be complete for one scope and not for another scope. The completeness of MSLE ontology can be assessed with SHACL and competency questions in domain-specific knowledge. Generally, there are two kinds of ontology completeness [24]. First, constraint-based completeness measures the percentage of concepts in an ontology that satisfy explicit representations of what must or must not be represented in the ontology. The constraints validate through some SHACL-SPARQL codes. Second, real-world-based completeness measures the degree to which certain real-world information is represented in the ontology. For instance, regarding detectors belonging to the dual beam electron microscope, calculating the completeness may consist in dividing "the number of detectors associated with this microscope in the ontology" by "the number of actual detectors associated with the dual beam."

## 2.7 Heavyweight Ontology of MSLE

After the evaluation stage using SHACL and SPARQL, the ontology can be improved to create a heavyweight version, which is more formal, more expressive, and consequently more reasoned. At this stage, more instantiations, axioms, properties, relationships, concepts, rules, and restrictions can be defined and modified to enhance the inference capability of the MSLE ontology and the semantics of the equipment. This, in turn, extends the knowledge coverage and expands the number of possible deductions.

## 3. Ontology development with Protégé

In this work, we used the Protégé Versions 5.5.0 [25] knowledge management system as an editor to develop the MSLE ontology. Protégé covers the most recent OWL 2 Web Ontology Language and RDF stipulation from the W3C association. Through protégé, it is possible to define and model the knowledge structure behind the data, as well as the reasoning and observations behind the knowledge structure itself. For example, in the MSLE ontology, we can derive new specifications or device categories from newly assigned equipment. The MSLE class hierarchy, visualization of the classes with VOWL (Visual Notation for OWL Ontologies), and the MSLE diagram are shown in Figure 10 and Figure 11.

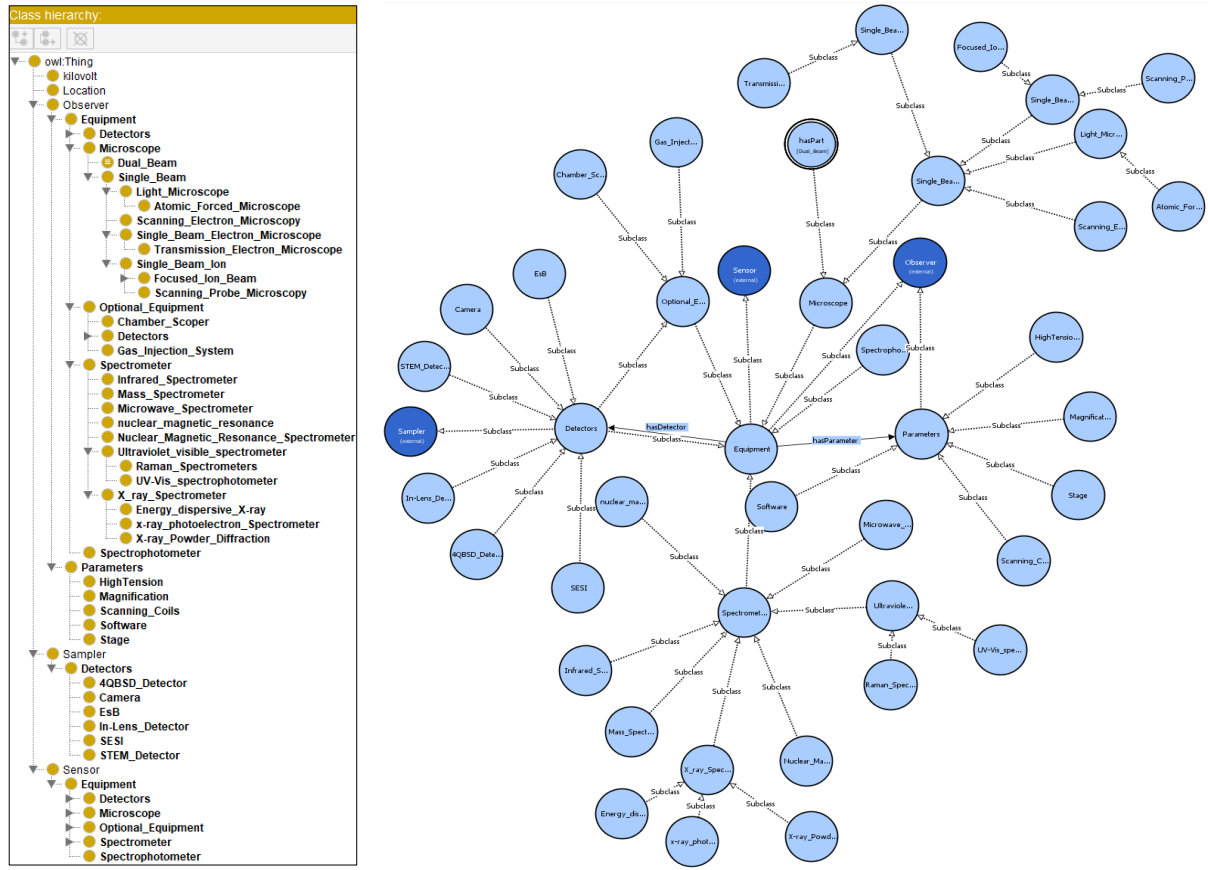

*Figure 10 The MSLE class hierarchy (left) and Ontology class visualization with VOWL (right) with Protégé*

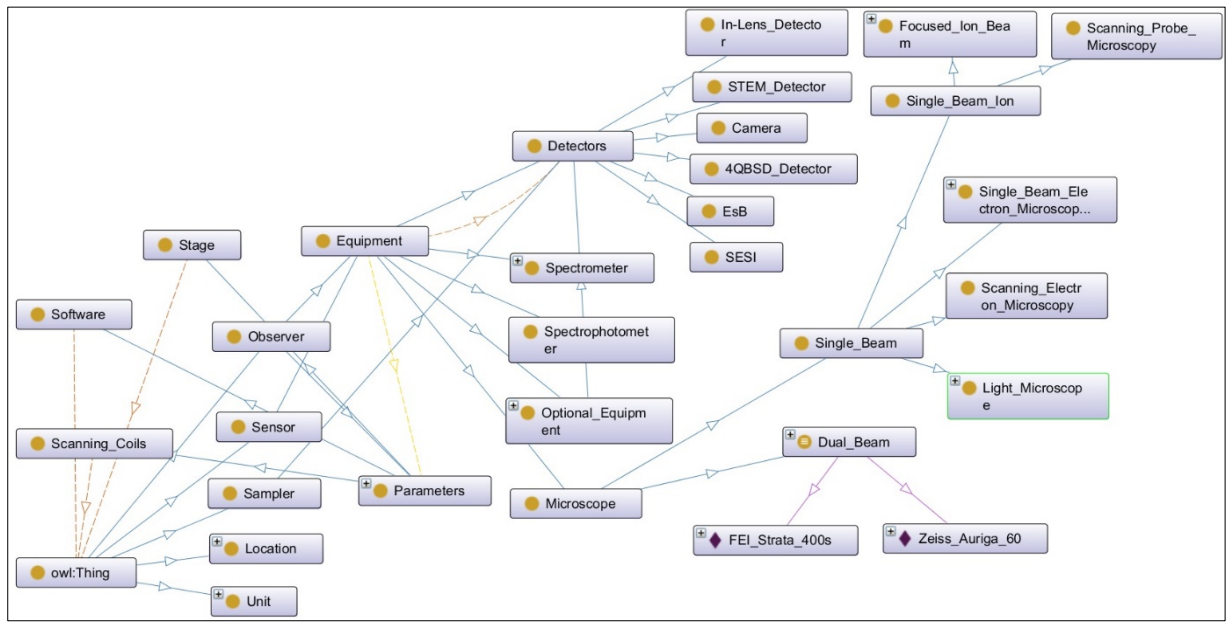

*Figure 11: The MSLE Ontology Diagram created by OntoGraf protégé*

## 4. Conclusions and Future Directions

This paper proposed the MSLE, an ontology for material science lab equipment, capable of efficiently representing the characterization equipment domain as basic semantic knowledge and guiding the user to document the equipment appropriately. SSN and MatVoc have been integrated into the MSLE core to build a coherent ontology and favor its interoperability. Terms related to the equipment were collected in various languages; abbreviations, acronyms, and terminologies were coded using the SKOS model. A list of competency questions provided by domain experts was used to develop the ontology. Furthermore, This work proposed a maturity model solution for assessing the MSLE ontology that consists of FAIR data principles, competency questions, SHACL, and completeness. All competency questions were verified to be consistent with the SPARQL language during the ontology assessment phase. Test cases based on existing knowledge in MSLE ontology have been carried out for demonstration purposes. The investigation of the maturity model defined on the ontology shows that the MSLE is a reliable tool to collect and exploit the knowledge about characterization lab equipment before and during experimental measurements. Due to its design, the MSLE has the potential to be further expanded to cover additional equipment in the future.

## Acknowledgments

The authors acknowledge funding by the Federal Ministry of Education and Research of Germany for the project STREAM ("Semantische Repräsentation, Vernetzung und Kuratierung von qualitätsgesicherten Materialdaten", ID: 16QK11C). This work has been supported by the Joint Lab "Integrated Model and Data Driven Materials Characterization" (JL MDMC), the research programs

"Engineering Digital Futures" and "Materials System Engineering" of the Helmholtz Association of German Research Centers and the Helmholtz Metadata Collaboration Platform. Furthermore, this project has received funding from the European Union's Horizon 2020 research and innovation programme under grant agreement No. 101007417 within the NFFA-Europe Pilot (NEP) Joint Activities framework.